%% file: Template.tex
\documentclass[9pt]{extarticle}
\usepackage{spconf,amsmath,graphicx,hyperref,amssymb}
\usepackage{booktabs}

\newenvironment{compactequation}{
  \setlength{\abovedisplayskip}{3pt}
  \setlength{\abovedisplayshortskip}{3pt}
  \setlength{\belowdisplayskip}{3pt}
  \setlength{\belowdisplayshortskip}{3pt}
  \begin{equation}
}{
  \end{equation}
  \ignorespacesafterend
}

\usepackage{caption}
\title{Reconstructing Topology-Consistent Face Mesh by Volume Rendering from Multi-View Images}
\name{Yating Wang$^1$, Ran Yi$^{1*}$, Xiaoning Lei$^{2}$, Ke Fan$^1$, Jinkun Hao$^1$, Lizhuang Ma$^{1*}$
\thanks{*Corresponding Authors}
}
\address{$^1$ Shanghai Jiao Tong University, $^2$ CATL}

\begin{document}
%
\maketitle
\begin{abstract}
Industrial 3D face assets creation typically reconstructs topology-consistent face meshes from multi-view images for downstream production. 
However, high-quality reconstruction usually requires manual processing or specific capture settings.
Recently NeRF has shown great advantages in 3D reconstruction, by representing scenes as density and radiance fields and utilizing neural volume rendering for novel view synthesis. 
Inspired by this, we introduce a novel method which combines explicit mesh with neural volume rendering to optimize geometry of an artist-made template face mesh from multi-view images while keeping the topology unchanged.
Our method derives density fields from meshes using distance fields as an intermediary and encodes radiance field in compact tri-planes. 
To improve convergence, several adaptions tailored for meshes are introduced to the volume rendering.
Experiments demonstrate that our method achieves superior reconstruction quality compared to previous approaches, validating the feasibility of integrating mesh and neural volume rendering.
\end{abstract}
\vspace{-5pt}
\section{Introduction}
\vspace{-5pt}

\label{sec:intro}
Reconstructing topology-consistent face meshes from multi-view images holds broad applicability in film production and VR/AR. 
While neural representations such as NeRF~\cite{nerf} and 3DGS~\cite{kerbl20233d} offer flexible geometry modeling, professional studios favor topology-consistent face meshes for their compatibility with animation rigs and real-time rendering. 
Moreover, face related applications often rely on topology-consistent face meshes, such as face prior models (e.g. BFM~\cite{paysan20093d}, FLAME\cite{li2017learning}, FaceScape\cite{yang2020facescape}), animatable head avatar\cite{wang20253d, Qian_2024_CVPR}.

\begin{figure}[t!]
    \centering
    \includegraphics[width=\linewidth]{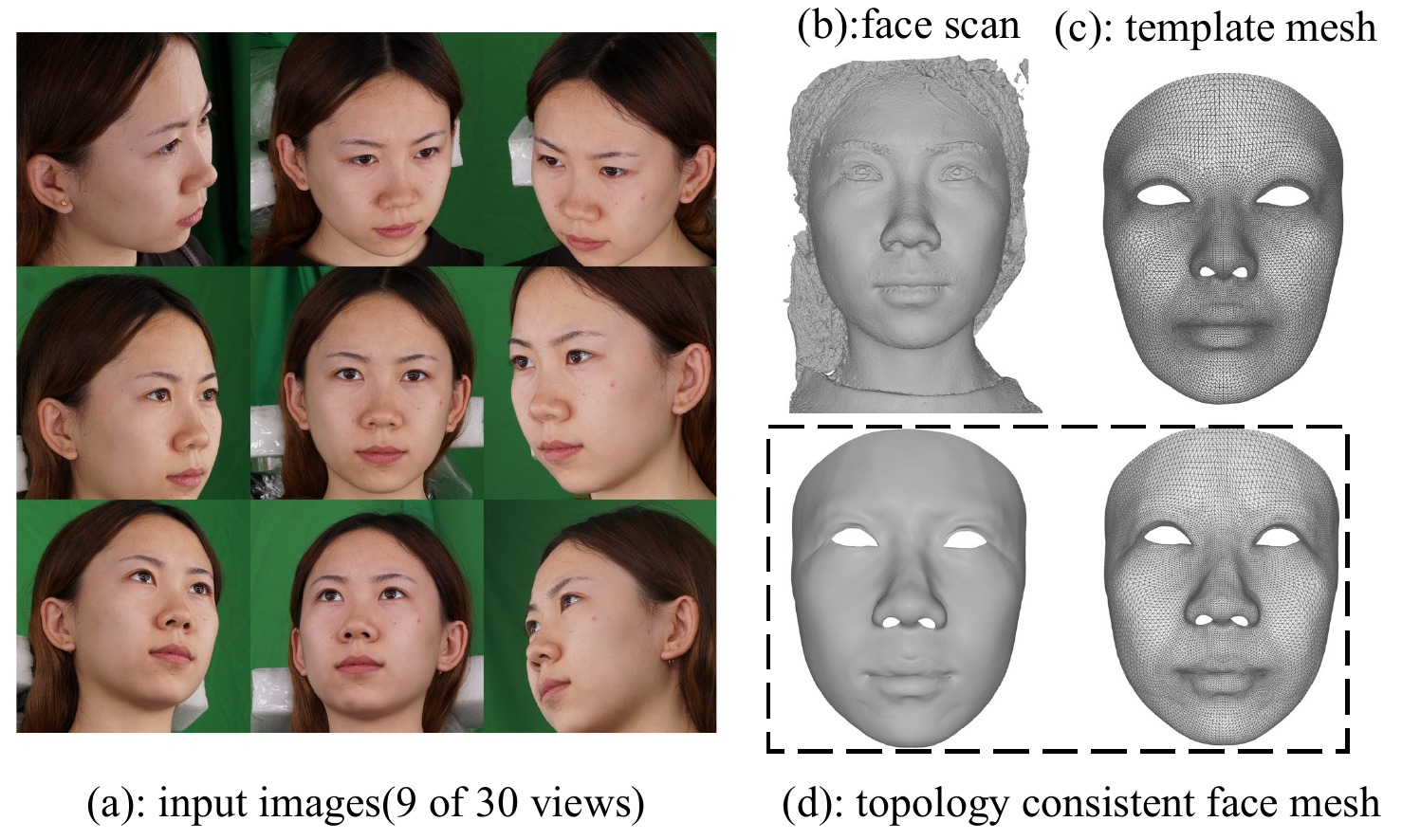}
    \caption{Our method takes multi-view images (a) as input and reconstructs face mesh (d) accurately while preserving the topology of the artist-designed template (c). We also show a raw face scan with arbitrary topology(b).}
    \label{fig:teaser}
    \vspace{-10pt}
\end{figure}

A classical two-step pipeline~\cite{3dmm, yang2020facescape} first reconstructs a raw shape with arbitrary topology (hereafter “scan”, as shown in Fig~\ref{fig:teaser}(b)) from multi-view images using photogrammetry tools (e.g., MetaShape, Colmap) or neural surface reconstruction methods (e.g., NeuS~\cite{neus}, NeUDF~\cite{23neudf}, SFDM~\cite{jin2025sfdm}). A template mesh(shown as Fig~\ref{fig:teaser}(c) ) is then non-rigidly registered to the scan via Non-Rigid Iterative Closest Point (N-ICP) to align topology. However, such scans are often noisy and require manual pre-processing to ensure robust N-ICP convergence.
Many end-to-end approaches have been proposed. 
VHAP~\cite{qian2024vhap} optimizes FLAME parameters and vertex offsets via neural rasterization. 
DFNRMVS~\cite{bai2020deep} combines inference with optimization for better generalization and learns person-specific blendshapes for lower error. Both methods rely on linear, low-rank shape priors for optimization robustness, which inherently constrain reconstruction accuracy.
ToFu~\cite{li2021tofu} and TEMPEH~\cite{bolkart2023instant}, directly regress topology-consistent meshes from multi-view inputs. While efficient, they demand large-scale ground-truth mesh datasets for training and exhibit limited generalization to images captured under different setups or alternative topologies.

Recent advances in 3D neural representations and rendering(e.g., NeRF~\cite{nerf},  NeuS~\cite{neus}, and Tri-planes~\cite{chan2022efficient}), achieve remarkable results in novel view synthesis and 3D reconstruction.
These methods store implicit density fields for geometry representation and view-dependent radiance fields for appearance, which are integrated into RGB colors through volume rendering, enabling end-to-end supervision from multi-view images to implicit fields. 
Inspired by these, we propose a novel end-to-end optimization-based topology-consistent face mesh reconstruction method, which simulates a density field through face mesh, encodes the radiance field by compact tri-planes and then optimize vertices positions of the artist-defined template face mesh with tri-planes parameters via neural volume rendering, while keeping the mesh topology unchanged.

Our key challenge is how to simulate the density field required for neural volume rendering using an explicit mesh. Inspired by neural implicit surface reconstruction methods such as NeuS~\cite{neus}, which derive density from signed distance fields via a well-designed mapping function for smoother surfaces but with arbitrary topology, we compute, for each sample point, its distance to the nearest triangle on the mesh and then map this distance to a corresponding density value.
We further propose several adaptions to neural volume rendering tailored for mesh properties to ensure the correctness and efficiency of the optimization.

We evaluate the proposed method on the multi-view face dataset introduced in \cite{relighting_zhibo}, benchmarking against representative reconstruction baselines under diverse settings. 
The experimental results validate the feasibility of combining mesh with volume rendering and show that the proposed method achieves better reconstruction accuracy, without face prior model and pre-training.
Moreover, our experiments show that the method remains robust even with sparse input views, such as only six views.
\vspace{-5pt}

\section{Method}
\vspace{-5pt}
\label{sec:method}
\input{method}

\vspace{-5pt}
\section{Experiments}
\vspace{-5pt}
\label{sec:exp}
\input{exp}

\vspace{-5pt}
\section{Conclusion.}
\vspace{-5pt}
We propose a novel method that integrates neural volume rendering with explicit mesh, enabling end-to-end topology-consistent face mesh reconstruction from multi-view images. Experiments demonstrate that this combination is feasible and produces high-quality reconstructions while preserving mesh topology — without relying on any face shape priors or pre-training. Moreover, our method generates plausible reconstructions even under sparse-view settings.

\bibliographystyle{IEEEbib}
\bibliography{strings,refs}

\end{document}

%% file: method.tex
\begin{figure}[t!]
    \centering
    \includegraphics[width=\linewidth]{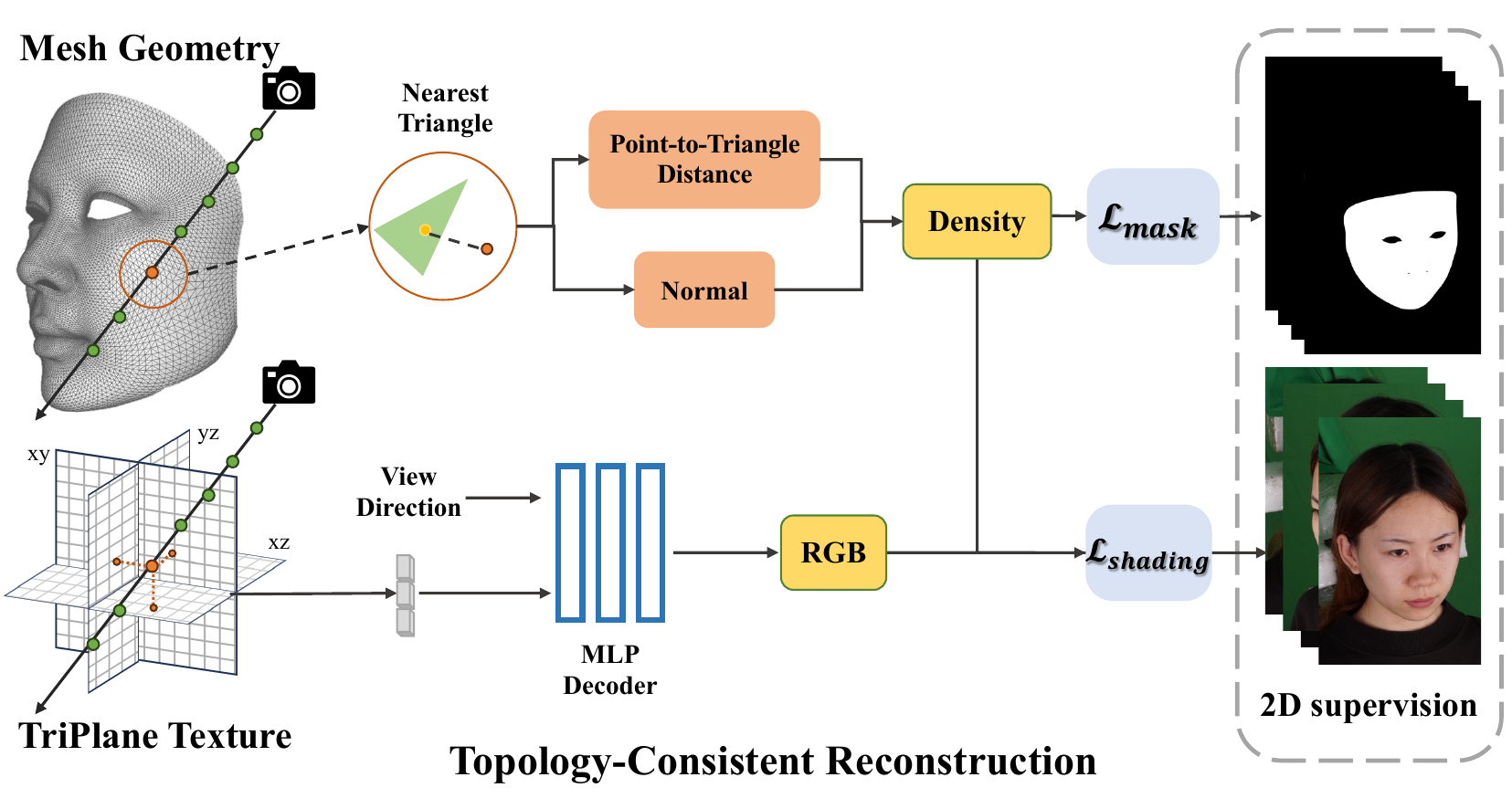}
    \caption{The pipeline of our method. 
    We introduce \textbf{mesh volume rendering} which transforms point-to-mesh distance into density, utilize tri-planes to store view-dependent face appearances, and generate high-quality images via volume rendering, thus enabling image supervision on the mesh geometry.}
    \label{fig:pipeline}
    \vspace{-10pt}
\end{figure}

Given multi-view images with masks and corresponding camera parameters, along with an artist-predefined face mesh template, we propose a novel \textbf{mesh volume rendering} mechanism that reconstructs a face mesh with fixed topology(the same as the mesh template).
Our pipeline is shown in Fig.~\ref{fig:pipeline}. 
\textbf{1)} In mesh volume rendering, for each sampled point along a ray, we search for the nearest point on the mesh surface. The distance between these two points is converted into density, and normal of the nearest triangle is assigned as normal of the sample point. 
\textbf{2)} For face appearance, we utilize axis-aligned tri-planes to store implicit encodings, and a tiny MLP decoder to decode the encodings along with view directions to RGB colors.
Based on this mesh volume rendering mechanism, we optimize vertices positions $V \in \mathbf{R}^{n_{v} \times 3}$ of a template mesh (whose topology $F \in \mathbf{Z}+^{n_t \times 3}$ is predefined by artists and fixed during optimization), along with tri-planes $T=\{T_{xy}, T_{xz}, T_{yz}\} \in \mathbb{R}^{3 \times n_f \times n_f \times n_{d}}$ and a tiny MLP decoder $D$ for face texture, 
to approximate multi-view images $\{I_i\}_{i=1}^{n_i}$ in $n_i$ views with known camera parameters (denoted by $\{K_i, R_i, t_i\}_{i=1}^{n_i}$) and masks $\{M_i\}_{i=1}^{n_i}$. 

\vspace{-5pt}
\subsection{Mesh Volume Rendering}
\vspace{-5pt}
\label{ssec:volume}
We introduce a neural volume rendering method for explicit meshes, adapting NeuS~\cite{neus}’s distance-to-density conversion with mesh-specific designs to ensure efficiency.
In NeuS~\cite{neus}, the geometry is represented by a \textbf{S}igned \textbf{D}istance \textbf{F}ield (SDF) rather than volume density in NeRF~\cite{nerf}. A mapping equation is designed to convert SDF values to volume rendering weights as shown in: 

\begin{compactequation}
    \alpha_{k} = max(1 - \frac{\Phi_{s}(s_{k+1})}{\Phi_{s}(s_{k})}, 0)
    \label{eq:sdf}
\end{compactequation}

where $s_k, \alpha_k$ is the SDF value and weight of the $k$-th sample point along $\mathbf{r}$, respectively. $\Phi_s$ is the cumulative distribution of logistic density distribution.
The accumulated color $\hat{C}$ of the ray $\mathbf{r}$ passing through a pixel is calculated by:

\begin{compactequation}
\hat{C} = \sum_k \prod_{j<k}(1-\alpha_j) \alpha_k \mathbf{c}_k
\label{eq:volume}
\end{compactequation}

SDF encodes the minimal distance from a spatial point to the surface. Given a mesh, we can compute the minimal distance from a  sample point $\mathbf{p}$ to the surface by finding its nearest point(noted as $\mathbf{p}'$) on the mesh, which may lie anywhere on the mesh surface, not necessarily at a vertex.
Following NeuS~\cite{neus}, we convert this distance into volume rendering weights via Eq.~\ref{eq:sdf}. However, NeuS is designed for watertight objects, whereas our target geometry — the facial region — is non-watertight. To handle open surfaces, we adopt the conversion from unsigned distance to volume rendering weights proposed in NeuDF~\cite{23neudf}, which extends NeuS to non-closed geometries.
For explicit distance fields, normals are computed via local derivatives. In NeuS, normals come directly from the MLP’s gradient. Meshes, however, only define normals on-surface via face cross-products. Computing off-surface normals by converting to an explicit field is costly. We find that approximating normal $\mathbf{n}$ of $\mathbf{p}$ with normal of $\mathbf{p}'$ is efficient and sufficiently accurate — especially since distant points contribute little in volume rendering.

Additionally, we employ back-face culling technique to mask out pixels covered only by triangles oriented away from the viewing direction.
To improve efficiency, we implement an OCTree in CUDA, replacing the naive nearest-point search (which iterates over all mesh triangles). The OCTree also enables us to identify and skip empty spatial regions during sampling, significantly accelerating the process.

\vspace{-5pt}
\subsection{Tri-planes Texture}
\vspace{-5pt}
\label{ssec:triplane}
The aforementioned mesh volume rendering allows simulating a density field by explicit mesh. We then employ compact tri-planes to store view-dependent radiance fields for face appearance modeling.
The tri-planes \(T\) consist of three orthogonal feature planes aligned with the axes: \(\{T_{xy}, T_{xz}, T_{yz}\} \in \mathbb{R}^{3 \times n_f \times n_f \times n_{d}}\), where $n_f\times n_f$ is plane resolution, and $n_{d}$ is the feature dimension. 
Observing that frontal views encode the majority of discriminative facial structure, we allocate higher dimension to $T_{xy}$ and lower to side views $T_{xz}, T_{yz}$.
The feature $t(\mathbf{p})$ of \(\mathbf{p}\) is obtained by projecting \(\mathbf{p}\) onto the $x-y, x-z$, and $y-z$ planes, 
bilinearly interpolating features on each plane, and concatenating the results. 
Then a tiny MLP $D$ decodes $t(\mathbf{p})$ and view direction $\mathbf{v}_c$ into RGB value $\mathbf{c}$. 
\textbf{P}osition \textbf{E}ncoding is employed on $\mathbf{v}_c$. This process can be formulated as:
\begin{compactequation}
    \mathbf{c} = D(\textbf{PE}(\mathbf{v}_c), t(\mathbf{p})).
\end{compactequation}

\vspace{-5pt}
\subsection{Loss and Progressive Training}
\vspace{-5pt}
\label{ssec:loss_training}
To reconstruct mesh with topology consistency from multi-view images based on our mesh volume rendering mechanism, we use the following five loss functions
to optimize 
vertices positions $V$, tri-planes $T$, and MLP decoder $D$
for geometry and appearance modeling: 

\noindent\textbf{Shading Loss.} We use $\mathcal{L}_1$ loss to constrain the shaded images $\hat{I}_i$ computed by mesh volume rendering being close to the real observations:
\begin{compactequation}
    \mathcal{L}_{color} = \sum_{i}^{n_i} |I_i - \hat{I}_i|.
    \label{eqn:shading_loss}
\end{compactequation}

\begin{figure}[t!]
    \centering
    \includegraphics[width=\linewidth]{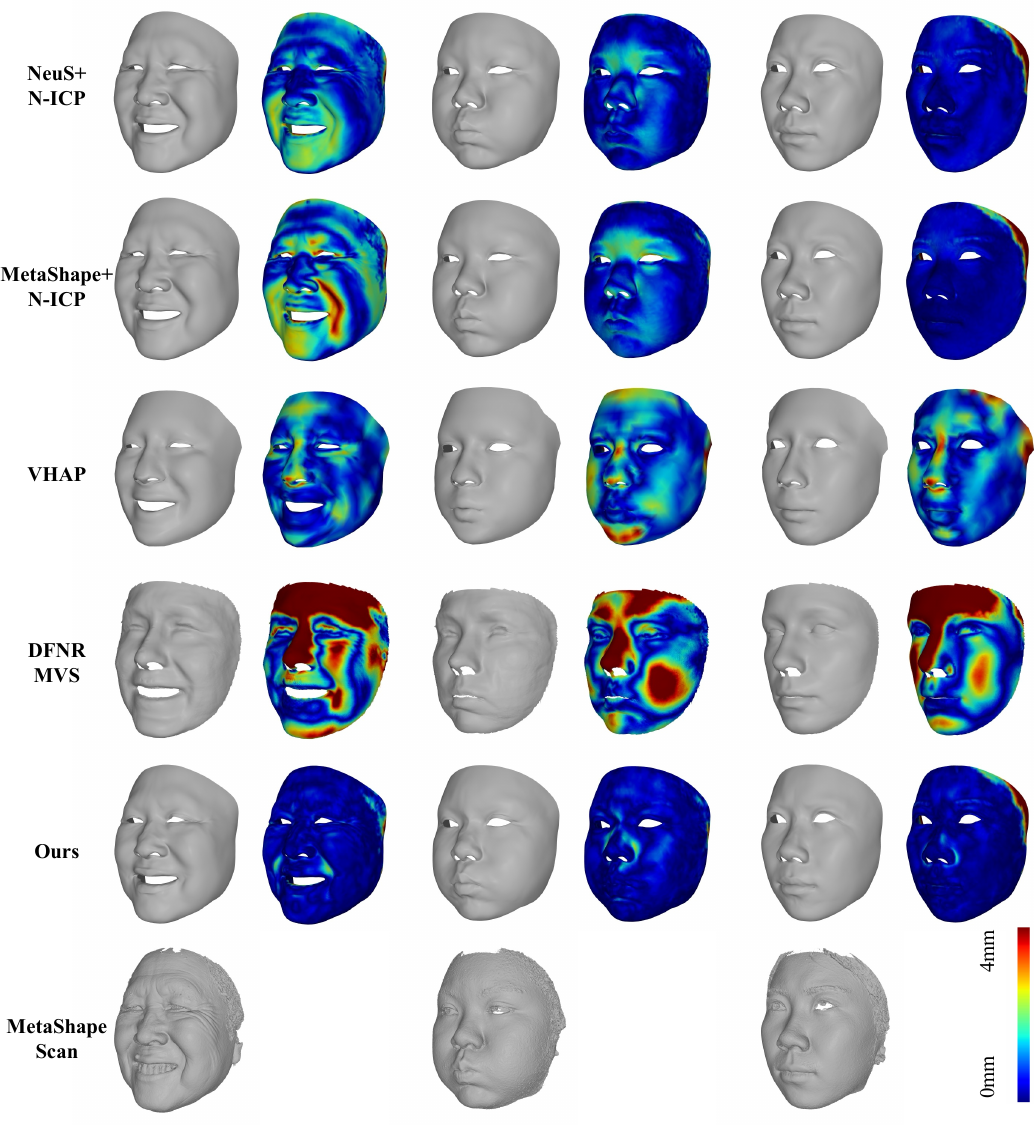}
    \caption{Qualitative comparison on reconstruction accuracy.}
    \label{fig:recon}
    \vspace{-10pt}
\end{figure}

\noindent\textbf{Tri-plane TV loss} is utilized to avoid noise in tri-planes and ensure stable training. We reduce the weight of TV loss as iteration increases. The TV loss is defined as the following equation, where $(u, v)$ denote a cell on a 2D plane:
\begin{compactequation}
\mathcal{L}_{\text{TV}} = \sum_{d \in \{xy, xz, yz\}} \sum_{u, v} \sqrt{(T_{u+1,v}^d - T_{u,v}^d)^2 + (T_{u,v+1}^d - T_{u,v}^d)^2}.
\end{compactequation}

\noindent\textbf{Landmark Loss} 
is involved to constrain that the projections of corresponding 3D vertices on the mesh are close to 2D detected landmarks during the initialization stage. 
We extract $n_l$ facial landmarks of selected front views denoted by $I_{f}$. The landmark loss can be formulated as: 
\begin{compactequation}
    \mathcal{L}_{ldmk} = \sum_{\mathcal{I}_f} \sum_{j=0}^{n_l} (\mathbf{p}_{j}^{p} - \mathbf{p}_{j}^{l})^2, 
    \quad
    \mathbf{p}_{j}^{p} = K(R\mathbf{v}_{j} + t).
\end{compactequation}
where $\mathbf{p}^p, \mathbf{p}^l$ are projection and detected landmarks, respectively. $K, [R|t]$ are camera intrinsic and extrinsic parameters for one view.

\noindent\textbf{Mask Loss.} The mask loss $\mathcal{L}_{mask}$ is defined as: 
\begin{compactequation}
\mathcal{L}_{mask} = ||M_k - \hat{O}_k||^2,
\end{compactequation}
where $\hat{O}_k = \sum_{i=1}^{n}T_{k,i}\alpha_{k,i}$ is the sum of weights along the camera ray. $M_k \in \{0, 1\}$ is the face mask for the $k$-th view. 
In practice, mask loss is computed solely on pixels around the mask contour to reduce sampling rays and accelerate training. 
We employ back culling to set $\hat{O}_k$ of pixels facing away from the camera to zero when calculating mask loss.

\noindent \textbf{Laplacian Loss.} 
Laplacian regularization~\cite{laplacian} promotes smooth mesh deformation by preserving local vertex-neighbor relationships. \cite{largestep} is a neural network compatible Laplacian regularization method with faster and more stable loss reduction. We use it as the shape regularization in our approach for stable convergence.
For a mesh with vertices 
$V=\{\mathbf{v_1},\mathbf{v_2},..., \mathbf{v_n}\}$, the Laplacian loss is defined as
\begin{compactequation}
    \mathcal{L}_{\text{lap}} = \sum_{i=1}^n \left\| \mathbf{\delta}_i \right\|^2, 
    \quad
    \mathbf{\delta}_i = \mathbf{v}_i - \frac{1}{|N(i)|} \sum_{j \in N(i)} \mathbf{v}_j
\end{compactequation}
where $N(i)$ is the set of neighboring vertices of $\mathbf{v_i}$ and $|N(i)|$ is the number of neighbors.

\noindent\textbf{Sampling.}
In NeuS~\cite{neus}, each iteration samples random 512 pixels from a randomly selected view. In experiments, we find that Laplacian regularization causes neighboring vertices to move together. To avoid oscillations, we use all training views per iteration. For shading loss, we sample pixels on a regular grid with small perturbations, balancing spatial regularity and sampling diversity.

\noindent \textbf{Progressive Training.}
A progressive training strategy is employed to ensure stable training and fast convergence. 
The first stage focuses on geometry initialization, where $\mathcal{L}_{ldmk}$ and $\mathcal{L}_{lap}$ are used to obtain initial guesses of vertices positions and then $\mathcal{L}_{mask}$ and $\mathcal{L}_{lap}$ are involved to optimize the initial guess to get a closer shape. 
In the second stage, based on the preliminary optimized geometry, we optimize the parameters of the tri-planes using $\mathcal{L}_{color}$ and $\mathcal{L}_{\text{TV}}$. 
In the final stage, we optimize both mesh geometry and the triplane textures using only $\mathcal{L}_{color}$ and $\mathcal{L}_{lap}$.
NeuS~\cite{neus} uses a learnable s to control density concentration. We fix s in early stages and optimize it in the final stage for better reconstruction.

%% file: exp.tex
\begin{figure}[t!]
    \centering
    \includegraphics[width=0.9\linewidth]{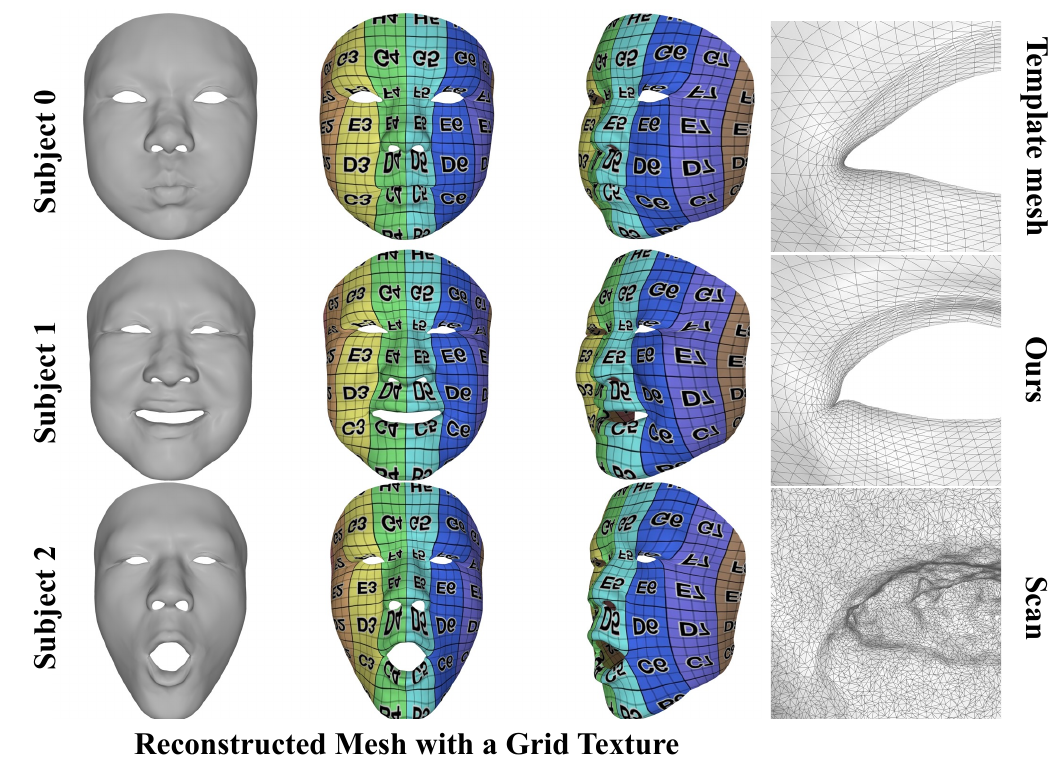}
    \caption{Qualitative analysis on topology consistency of the reconstructed meshes.}
    \label{fig:topo}
    \vspace{-10pt}
\end{figure}

\begin{table}[t!]
    \centering
    \fontsize{9pt}{11pt}\selectfont
    \resizebox{0.75\linewidth}{!}{
    \begin{tabular}{c|c|c}
    \toprule
        Method & Error $\downarrow$ & Vertex Num \\
    \midrule
        Metashape& - & 110,340\\
        Metashape + N-ICP&3.725&11,763\\
        NeuS + N-ICP & 4.412 & 11,763 \\
        VHAP~\cite{qian2024vhap} & 14.21 & 1787\\
        DFNRMVS~\cite{bai2020deep} & 18.95 & 28,854\\
        Ours&3.255&11,763 \\
    \bottomrule
    \end{tabular}
    }
    \caption{Quantitative comparison of reconstruction accuracy. }
    \label{tab:geo}
    \vspace{-10pt} 
\end{table}

\vspace{-5pt}
\subsection{Implementation Details}
\vspace{-5pt}
We use artist-designed Unreal MetaHuman mesh as template. Following 3DMM~\cite{3dmm}, we manually extract face region (11,763 verts, 23,004 tris) and label 106 landmarks. We conduct evaluations on the dataset from~\cite{relighting_zhibo} which includes 30-view 5456×3632 face images and landmark annotations. Camera parameters and scans are estimated using Agisoft Metashape and manually refined as ground truth. We obtained 7 authorized scans of different expressions from 4 people. Facial masks are generated by projecting the registered mesh using Metashape cameras, or alternatively via face parsing~\cite{yu2018bisenet}. 
Our method is implemented in PyTorch and trained on a single NVIDIA A6000 GPU using the Adam optimizer. 
We employ a three-layer MLP with ReLU activation as the tri-plane decoder, with feature dimension $n_f = 64$. 
The plane-specific dimensions are set to $n_d = 32$ for $T_{xy}$ and $n_d = 16$ for $T_{xz}$ and $T_{yz}$. 
Learning rates are $2 \times 10^{-3}$ for tri-plane features, $5 \times 10^{-4}$ for the decoder, and $3 \times 10^{-2}$ for mesh vertices. 
We set the Laplacian regularization weight to 19 and exclude face contour landmarks due to semantic inconsistency across views. 
Training proceeds in three stages: 
(1) 40 iterations of $\mathcal{L}_{\text{ldmk}}, \mathcal{L}_{\text{mask}}$ for geometry initialization; 
(2) 300 epochs of $\mathcal{L}_{\text{color}}, \mathcal{L}_{\text{TV}}$ with fixed geometry; 
(3) 300 epochs of joint optimization of geometry and tri-planes.

\vspace{-5pt}
\subsection{Comparisons.}
\vspace{-5pt}

\begin{figure}[t!]
    \centering
    \includegraphics[width=0.7\linewidth]{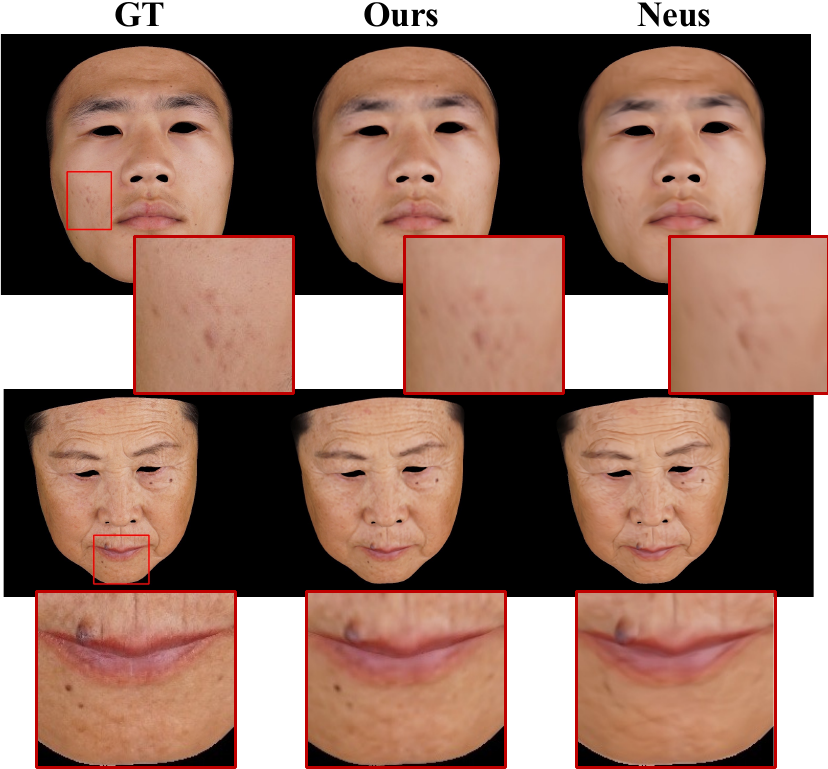}
    \caption{Qualitative comparison with NeuS on rendering quality.}
    \label{fig:render}
    \vspace{-10pt}
\end{figure}

\begin{table}[t!]
    \centering
    \fontsize{9pt}{11pt}\selectfont
    \resizebox{0.7\linewidth}{!}{
    \begin{tabular}{c|c|c|c}
        \toprule
         Method& PSNR$\uparrow$&SSIM$\uparrow$&LPIPS$\downarrow$\\
         \midrule
         NeuS&  30.2021&0.8114&0.3399\\
         Ours& 32.1003&0.9183&0.09155 \\
         \bottomrule
    \end{tabular}
    }
    \caption{Quantitative comparison with NeuS on rendering quality.}
    \vspace{-10pt} 
    \label{tab:render}
\end{table}

\noindent\textbf{Reconstruction Accuracy.}
Face scans generated by Agisoft Metashape(a commercial MVS software based on image feature matching) are used as shape ground truth.
The average nearest distance per vertex between reconstruction and Metashape scans is calculated for quantitative comparison.
We compare with the following baselines, the quantitative results are shown in Tab~\ref{tab:geo}, and error per vertex ismapped to color for visualization in Fig~\ref{fig:recon}. 
1) Classical two step pipeline that employs N-ICP registration on Metashape scans or NeuS reconstruction.
2) VHAP, a common used tool to reconstruct FLAME face mesh from multi-view images.
3) DFNRMVS, which leverages BFM shape prior to reconstruct face mesh from non-rigid multi-view images.
The hair region with relatively high noise is excluded when calculating distances.

\noindent\textbf{Topology Consistency.}
We demonstrate that our method maintains topology consistency across various facial expressions and identities by applying a grid texture, which can be found in Fig~\ref{fig:topo}.
We also show zoomed-in views of the eye region on template face mesh, our reconstruction and scan.
Our result faithfully retains the artist-crated topology of template mesh, which features a semantically optimized vertex distribution.

\noindent\textbf{Rendering Quality.}
We compare our method with NeuS on rendering quality to validate the feasibility of combining explicit mesh and neural voluem rendering.
We select one held-out view for testing from 30 viewpoints. PSNR, SSIM, LPIPS~\cite{zhang2018unreasonable} are computed to evaluate differences between rendered images with true observations, which can be found in Table~\ref{tab:render}. Our method can render facial details as shown in the zoom-in view of Fig~\ref{fig:render}.

\vspace{-5pt}
\subsection{Ablation Study.}
\vspace{-5pt}
\begin{figure}[t!]
    \centering
    \includegraphics[width=0.8\linewidth]{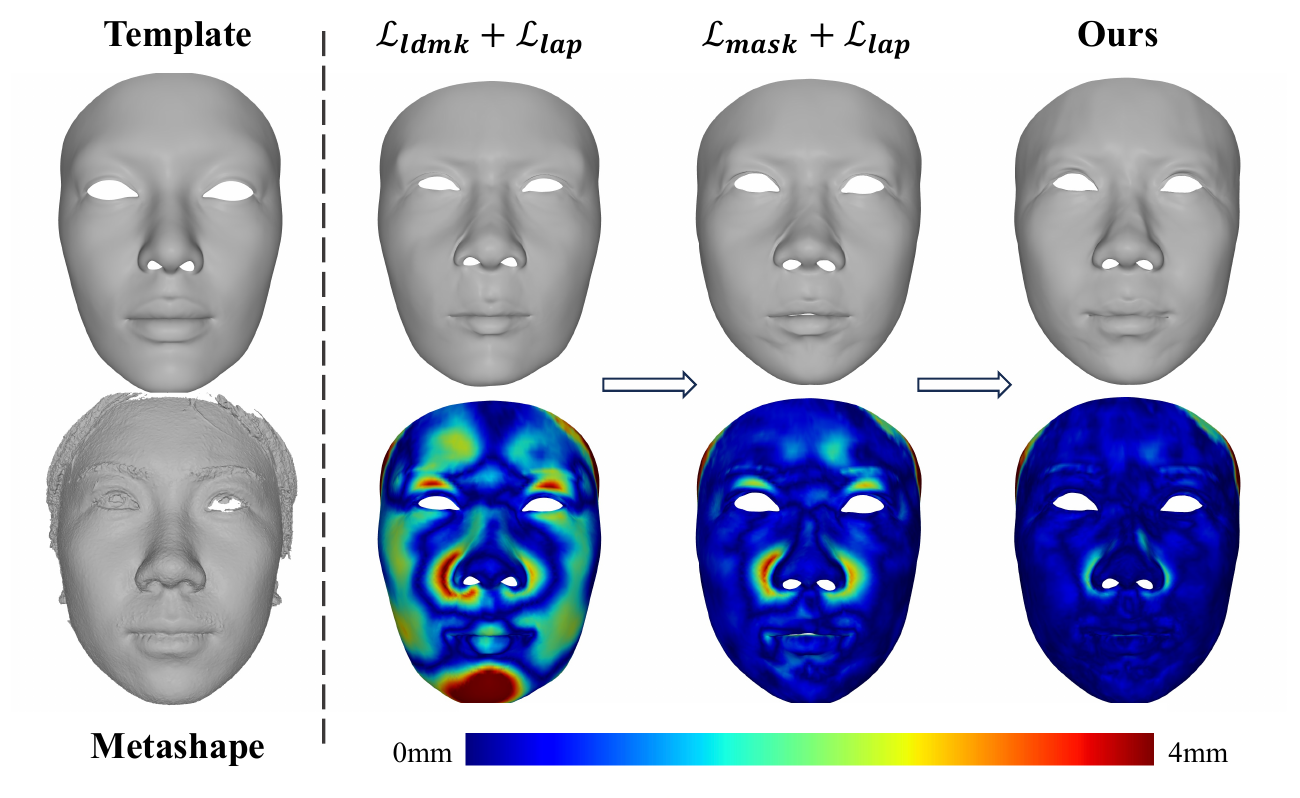}
    \caption{Ablation Study on Progressive Training.}
    \label{fig:training}
    \vspace{-10pt}
\end{figure}
\noindent\textbf{Progressive Training.}
To demonstrate the results of progressive training, we show in Fig.~\ref{fig:training} 
1) face geometry only using $\mathcal{L}_{ldmk}$ and laplacian rigidity constraints, 
2) finetuning 1) with $\mathcal{L}_{mask}$, 
3) final reconstruction using $\mathcal{L}_{color}+\mathcal{L}_{lap}$. An approximate shape is obtained through optimization with sparse landmarks and mask constraints, whereas the dense shading loss enables precise reconstruction by providing detailed photometric supervision.

\begin{figure}[t!]
    \centering
    \includegraphics[width=0.8\linewidth]{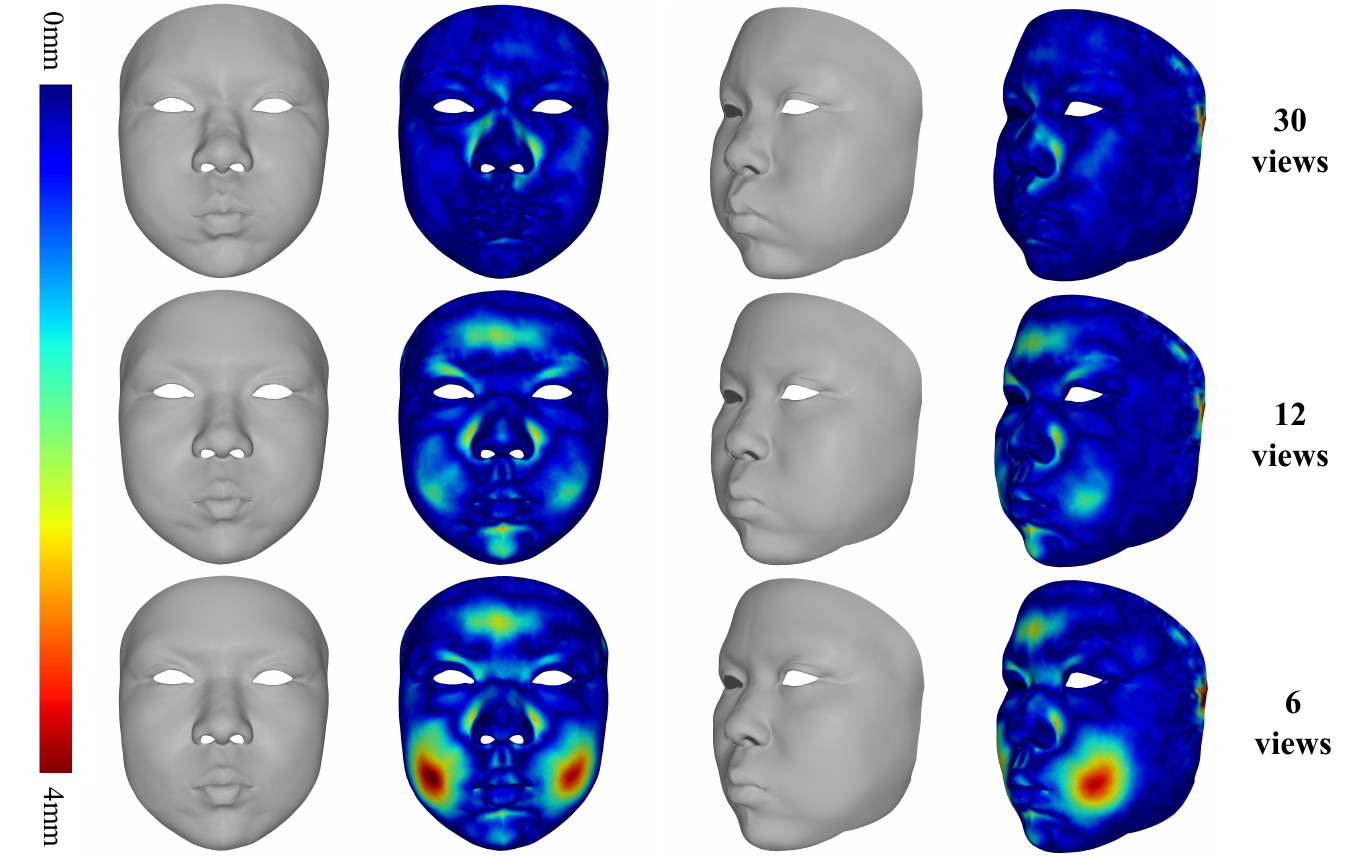}
    \caption{Ablation study on Input Views.}
    \label{fig:views}
    \vspace{-10pt}
\end{figure}

\begin{table}[t!]
    \centering
    \fontsize{9pt}{11pt}\selectfont
    \begin{tabular}{c|c|c|c}
        \toprule
         Views& 30 & 12 & 6\\
         \midrule
         Geometry Error& 3.255& 4.589& 5.901\\
         \bottomrule
    \end{tabular}
    \caption{Reconstruction accuracy of different input views.}
    \label{tab:views}
    \vspace{-10pt}
\end{table}

\noindent\textbf{Input Views}
We present our results under different input views to verify the robustness of our method.
From all 30 input views, we select the 12 most frontal and 6 most frontal views. As input views decreases, the reconstruction accuracy gradually degrades, as shown in Tab~\ref{tab:views}. 
However, as our method optimizes geometry based on the artist-defined mesh, even with only 6 input views, our method is still able to produce reasonable reconstructions (as shown in Fig~\ref{fig:views}).